\documentclass[lettersize,journal]{IEEEtran}
\usepackage{amsmath,amsfonts}
\usepackage{algorithmic}
\usepackage{algorithm}
\usepackage{array}
\usepackage[caption=false,font=normalsize,labelfont=sf,textfont=sf]{subfig}
\usepackage{textcomp}
\usepackage{stfloats}
\usepackage{url}
\usepackage{verbatim}
\usepackage{graphicx}
\usepackage{cite}
\usepackage{color}
\usepackage{booktabs}
\usepackage{multirow}
\usepackage{makecell}
\usepackage{float}
\usepackage{cleveref}

\begin{document}

\title{CtxMIM: Context-Enhanced Masked Image Modeling for Remote Sensing Image Understanding}

\author{Mingming Zhang, Qingjie Liu,~\IEEEmembership{Member,~IEEE,} and Yunhong Wang,~\IEEEmembership{Fellow,~IEEE}
\thanks{Mingming Zhang, Qingjie Liu, and Yunhong Wang are with the State Key Laboratory of Virtual Reality Technology and Systems, Beihang University, Beijing 100191, China, and also with the Hangzhou Innovation Institute, Beihang University, Hangzhou 310051, China (e-mail: sara\_@buaa.edu.cn; qingjie.liu@buaa.edu.cn; yhwang@buaa.edu.cn).}
}

\maketitle

\begin{abstract}
Learning representations through self-supervision on unlabeled data has proven highly effective for understanding diverse images. However, remote sensing images often have complex and densely populated scenes with multiple land objects and no clear foreground objects. This intrinsic property generates high object density, resulting in false positive pairs or missing contextual information in self-supervised learning. To address these problems, we propose a context-enhanced masked image modeling method (CtxMIM), a simple yet efficient MIM-based self-supervised learning for remote sensing image understanding. CtxMIM formulates original image patches as a reconstructive template and employs a Siamese framework to operate on two sets of image patches. A context-enhanced generative branch is introduced to provide contextual information through context consistency constraints in the reconstruction. With the simple and elegant design, CtxMIM encourages the pretraining model to learn object-level or pixel-level features on a large-scale dataset without specific temporal or geographical constraints. Finally, extensive experiments show that features learned by CtxMIM outperform fully supervised and state-of-the-art self-supervised learning methods on various downstream tasks, including land cover classification, semantic segmentation, object detection, and instance segmentation. These results demonstrate that CtxMIM learns impressive remote sensing representations with high generalization and transferability.
\end{abstract}


\section{Introduction}
\label{sec:intro}

\IEEEPARstart{R}{emote} sensing images have been widely used in many applications, including precision agriculture, disaster management, city planning, and environment monitoring. With a growing number of satellites collecting remote sensing images ceaselessly, abundant remote sensing images are easy to access. However, annotations are technically challenging due to highly time-consuming and expertise knowledge than general benchmark datasets, \textit{e.g.}, ImageNet \cite{r2015imagenet}. Recently, self-supervised learning (SSL) has been increasingly emerging to pretrain models on extensive unlabeled data and demonstrated more robust performance than supervised learning on classification, segmentation, and detection. Therefore, the SSL paradigm has been receiving much attention from the remote sensing community and exploring to obtain an efficient pretrained model on large-scale unlabeled data for remote sensing image understanding.

\begin{figure} [tb]
\centering
\includegraphics[width=\linewidth,scale=1.0]{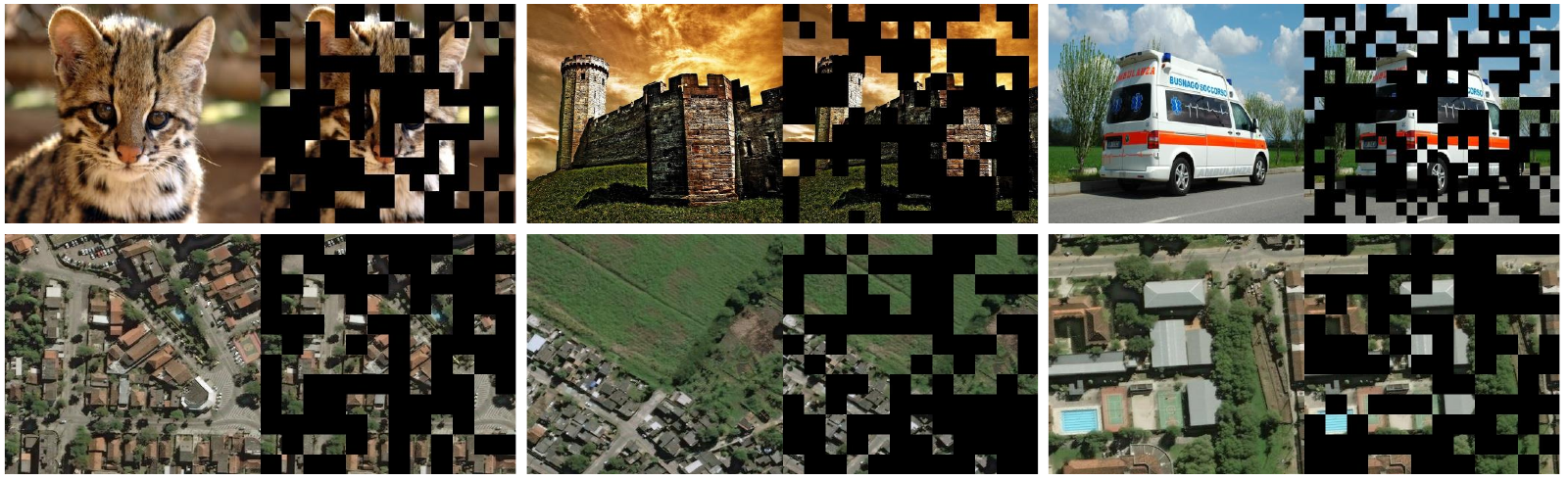}
\caption{Object density gap between natural and remote sensing images. Natural images have apparent foreground objects in a relatively simple background, but remote sensing images contain multiple objects, especially no apparent foreground objects, in a vast and complicated scene. Hence, extending MIM to remote sensing images will miss contextual information during reconstruction due to considerably high object density.}
\label{fig:objden}
\end{figure}

\begin{figure*} [htp]
\centering
\includegraphics[width=\linewidth,scale=1.0]{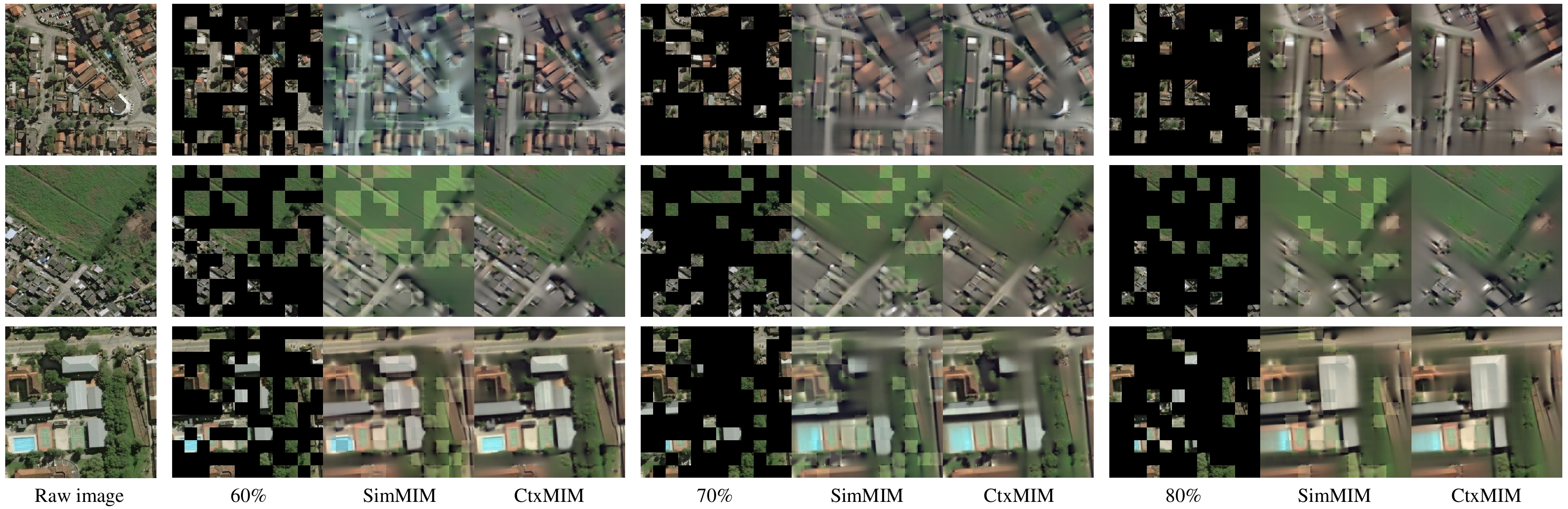}
\caption{Reconstructive results on remote sensing images. The first column is raw images. The other triples show the masked image with different masking ratios (left), SimMIM \cite{xie2022simmim} reconstruction (middle), and our CtxMIM reconstruction (right). SimMIM \cite{xie2022simmim} and the proposed CtxMIM are pretrained on our collected dataset. Although more and more land objects are masked as the ratio becomes larger, our CtxMIM can still reconstruct masked images better in terms of texture and content.}
\label{fig:misscontext}
\end{figure*}

Existing advanced SSL methods \cite{m2021seco,a2021geoaware,mall2023caco} primarily adopt contrastive learning \cite{c2020mocov2,c2020simclr,c2021simsiam} for self-supervised remote sensing representation learning. Indeed, contrastive approaches learn representations invariant to season, geographical location, illumination, and viewing angle by employing extensive data augmentation, which have exceeded supervised learning-based methods. However, state-of-the-art contrastive methods typically require careful pretraining data organization in terms of temporal and geographical dimensions to prevent the pretraining collapse. Additionally, these methods generally concentrate on image-level feature learning by comparing two augmented views.

Recently, SSL based on masked image modeling (MIM) has demonstrated potential performance in natural image representation learning \cite{he2022mae,xie2022simmim}. These MIM methods reconstruct masked patches by randomly masking an image with a high masking ratio (\textit{e.g.}, 75\% in MAE \cite{he2022mae} and 80\% in SimMIM \cite{xie2022simmim}), which can learn good representations without extensive data augmentation. The pixel reconstruction idea of MIM is well-suited for remote sensing tasks since remote sensing analysis typically requires object-level or pixel-level features with fine-grained details rather than image-level features. However, applying MIM from natural images to remote sensing images still has limitations due to the considerably high object density in remote sensing images. Natural images have apparent foreground objects in a relatively simple background, \textit{e.g.}, the cat, castle, and ambulance at the top of Figure \ref{fig:objden}, which can be easily repaired by inferring the whole from some parts through rich contextual information. Since remote sensing images are captured by sensors mounted on satellites, they generally contain multiple objects in a vast and complicated scene and typically have no apparent foreground objects, as illustrated at the bottom of Figure \ref{fig:objden}. Unlike natural images, some small land objects may be entirely masked by randomly masking, making pretraining convergence difficult, as demonstrated in ablation studies of subsection \ref{subsec:ablation}. Hence, MIM methods with a high masking ratio would hamper semantic extrapolating for remote sensing representation learning due to missing contextual information, as shown in Figure \ref{fig:misscontext}.

To address these issues, we propose a novel \textbf{\underline{C}}on\textbf{\underline{t}}e\textbf{\underline{x}}t-enhanced \textbf{\underline{M}}asked \textbf{\underline{I}}mage \textbf{\underline{M}}odeling method (CtxMIM) to learn representations with high-level semantic and fine-grained details for remote sensing image analysis. CtxMIM first adopts a pixel reconstructive pre-text task to capture local information from the masked image. Then, a context-enhanced generative branch formulates the original patches as the reconstructive template, encouraging the reconstruction learning semantic extrapolating by providing meaningful contextual information. To train CtxMIM, a large-scale unlabeled dataset is collected, which contains over 1.28 million images without being carefully designed in terms of temporal and geographical dimensions.

We evaluate our approach on different tasks, including land cover classification, semantic segmentation, object detection, and instance segmentation. Following SOTA remote sensing SSL methods, we use EuroSAT \cite{helber2019eurosat}, NWPU-RESISC45 \cite{cheng2017nwpu}, DOTA \cite{xia2018dota}, and SpaceNet \cite{van2018spacenet} for comparison. Extensive experiments demonstrate that our method outperforms SOTA SSL methods and is more effective for remote sensing tasks than supervised learning on the common ImageNet \cite{r2015imagenet}. In summary, our contributions are:
\begin{enumerate}
    \item We propose a novel context-enhanced masked image modeling method (CtxMIM), a simple self-supervised learning framework to learn robust and transferable representations for efficient remote sensing image analysis.  
    \item CtxMIM is based on the analysis that remote sensing images have considerably higher object density than natural images. CtxMIM formulates the original image patches as the reconstructive template and introduces a context-enhanced generative branch to provide contextual information.
    \item By pretraining on a large-scale unlabeled dataset, CtxMIM learns good representations to benefit image-level, object-level, and pixel-level downstream tasks. Moreover, we achieve state-of-the-art performance on land cover classification, semantic segmentation, object detection, and instance segmentation, demonstrating its transferability and superiority for remote sensing image analysis.
\end{enumerate}

The remainder of this paper is organized as follows: Section \ref{sec:related} reviews related studies. Section \ref{sec:method} introduces our proposed CtxMIM in detail. Section \ref{sec:exp} first describes experimental settings, comparison methods, and evaluation metrics, then reports and discusses the experimental results quantitatively and qualitatively. Section \ref{sec:dis} analyzes remote sensing pretraining in terms of model structure, performance, limitations, and future work. Section \ref{sec:con} concludes this work.

\section{Related work}
\label{sec:related}
In this section, we first introduce two dominant self-supervised learning frameworks (\textit{i.e.}, contrastive learning, and reconstructive learning) in the computer vision field. Motivated by increasing success in many computer vision tasks, self-supervised learning has been a prevalent paradigm for remote sensing pretraining, which is discussed in the following \ref{subsec:sslrs}.

\subsection{Contrastive Learning}
\label{subsec:cl}
Contrastive learning based methods \cite{kermiche2019contrastive,t2020cmc} learn good representations by minimizing the distance between representations of positive pairs augmented from one image and pushing representations of negative pairs from different images away. SimCLR \cite{c2020simclr} systematically analyzes the principles of contrastive learning methods and designs a simple framework for representation learning based on a Siamese network and extensive data augmentations. MoCos \cite{he2020mocov1,c2020mocov2} introduce a dynamic large queue from different batches for contrastive learning and employ a moving-averaged encoder to maintain consistency. Since these methods require large batches or negative pairs to avoid collapsing, some methods \cite{g2020bootstrap,c2020SwAV,c2021simsiam,z2021barlow} redesign the network structures, propose an innovative objective function, or introduce online clustering for representation learning. Recent contrastive methods \cite{liu2021anomaly,zhai2022mvcnet,wu2023practical,yan2024hcl} have demonstrated superiority in different downstream tasks, even outperforming supervised learning. Although increasingly successful in many computer vision tasks, contrastive methods rely on extensive data augmentations and focus on image-level features, limiting the transferring performance in object-level or pixel-level cases.
 
\subsection{Reconstructive Learning}
\label{subsec:rl}
Based on the idea of masked language modeling in NLP, reconstructive learning \cite{he2022mae,xie2022simmim,wei2022maskfeat} learns representations through solving the reconstructive pre-text task. This line of work randomly masks a portion of image patches using a special [MASK] token and then predicts raw pixel values using masked image modeling. SimMIM \cite{xie2022simmim} only uses a lightweight prediction head to reconstruct raw pixel values of masked patches, facilitating good representation learning of the encoder. MAE \cite{he2022mae} first handles visible patches in the encoder, and then encoded patches and mask tokens are processed in the decoder, which can significantly reduce the computation overhead of the encoder. MixMAE \cite{liu2023mixmae} reconstructs two original images from the mixed image by replacing the original masked patches with visible patches of another image, improving pretraining efficiency and consistency. With SOTA performance on various tasks, many related works have been studied, such as VideoMAE \cite{t2022videomae}, MultiMAE \cite{b2022multimae}, DMDSC \cite{yang2023dual}, MCGMAE \cite{fu2024multilevel}, and SiamMAE \cite{gupta2024siamese}.

\subsection{Self-Supervised Learning in Remote Sensing}
\label{subsec:sslrs}
Recently, self-supervised learning has been broadly applied in remote sensing domains. For instance, many methods \cite{k2020deep,v2021color,l2021semantic,s2021cmc,j2021contrastive} utilize contrastive learning to learn representations.
SeCo \cite{m2021seco} contrastively learns seasonal invariance by constructing a remote sensing dataset of different temporal information. Some methods \cite{a2021geoaware,l2021geoknowledgel,l2022geographical} consider that geographical knowledge is the unique characteristic of remote sensing images, which can be used for contrastive learning. GeRSP \cite{huang2024gersp} simultaneously learns representations with general and special knowledge from remote sensing and natural images by a teacher-student architecture. It designs self-supervised and supervised pretraining branches to generate the powerful pretrained model for remote sensing image understanding.
Another attempt \cite{c2022satmae,mendieta2023gfm,qi2024masked,hong2024spectralgpt} explores the masked image modeling paradigm for remote sensing feature pretraining by repairing masked images.
SatMAE++ \cite{noman2024satmaepp} leverages the scale information of remote sensing imagery to pretrain a transformer framework, finally achieving representations with more scales.
ScaleMAE \cite{reed2023scalemae} reconsiders the scale information by explicitly learning relationships between data at different scales. After encoding a masked image, Scale-MAE reconstructs low/high frequency images at lower/higher scales.
CrossScaleMAE \cite{tang2024csmae}, based on the Masked Auto-Encoder (MAE), adopts scale augmentation techniques and introduces cross-scale consistency constraints to pretrain consistent and meaningful representations for downstream tasks.
Besides, more relevant works have been studied for particular tasks, including change detection \cite{wu2021com,luppino2022code,qu2024cycle}, land cover classification \cite{liu2022contrastive,kalita2022class,aksoy2022multi}, and semantic segmentation \cite{l2022gl,m2022index,tang2023semantic}.

While many methods have improved performance on different tasks, these methods are pretrained on datasets by carefully organizing in terms of temporal and geographical dimensions, leading to poor generalization ability. Since remote sensing images contain multiple interesting objects rather than one obvious foreground object in natural images, SSL methods remain challenging in the acquisition of positive pairs and reconstruction of masked image patches due to high object density. This work designs self-supervised learning based on reconstruction with a context-enhanced generative branch, which can learn representations with rich semantic and fine-grained details.

\section{Method}
\label{sec:method}
In this section, We introduce the proposed CtxMIM, a novel MIM-based self-supervised learning for remote sensing image understanding. In the following, we will first describe the pretraining pipeline of the proposed CtxMIM. Then, we describe the main components of CtxMIM.

\begin{figure} [tp]
\centering
\includegraphics[width=\linewidth, scale=1.00]{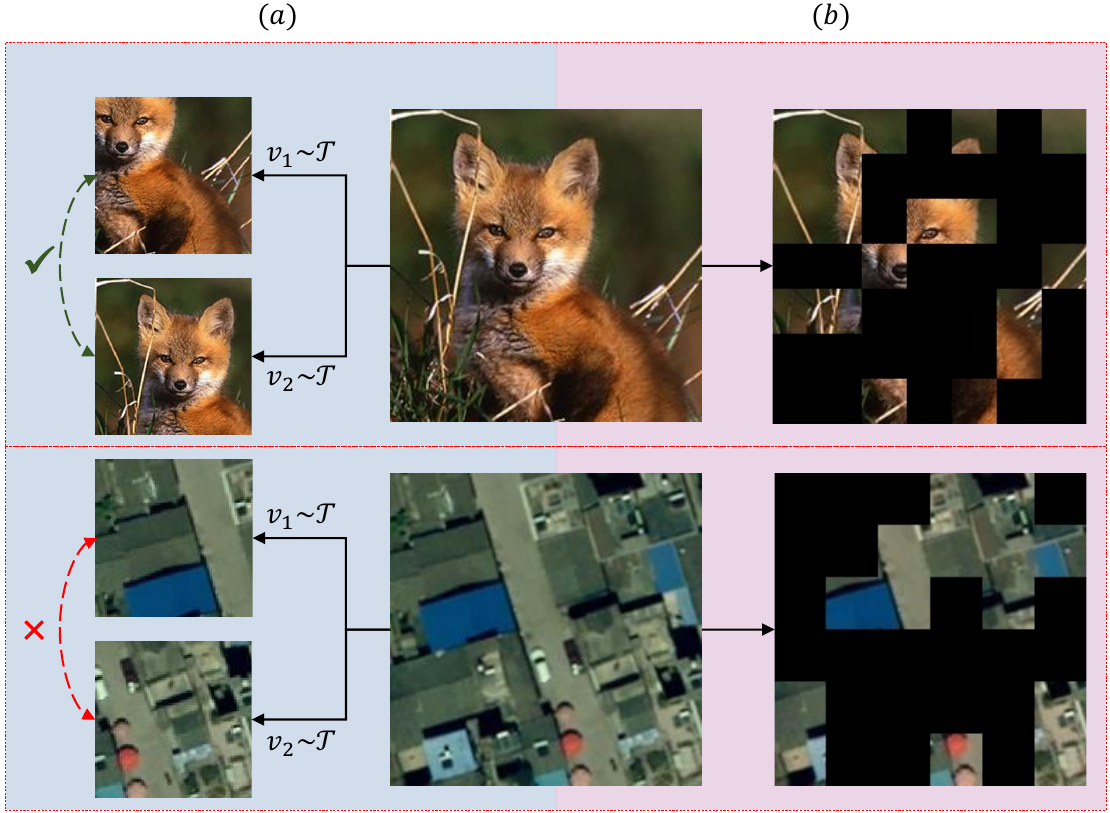}
\caption{High object density in remote sensing images, resulting in mismatched positive pairs in contrastive learning or missing contextual information in reconstructive learning.}
\label{fig:bg2M}
\end{figure}

\begin{figure*} [tbp]
\centering
\includegraphics[width=\linewidth,scale=1.00]{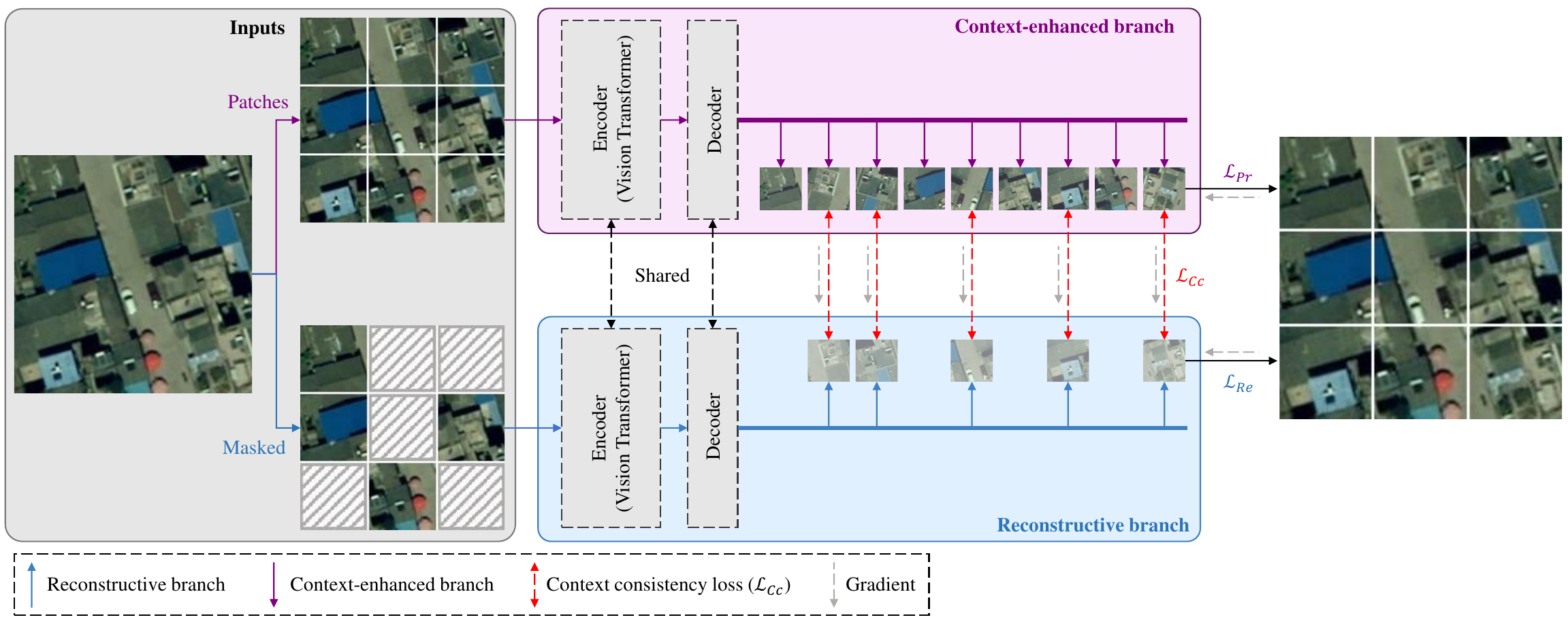}
\caption{An illustration of CtxMIM, a simple yet efficient pretraining framework for remote sensing tasks. CtxMIM introduces a novel context-enhanced generative branch to provide contextual information by the context consistency constraint (${\mathcal L}_{Cc}$) during the reconstruction, which formulates the original image patches as the reconstructive template. CtxMIM learns highly generalizable and transferable representations for various downstream tasks (\textit{e.g., image-level, object-level, and pixel-level}).}
\label{fig:overview}
\end{figure*}

\subsection{Overview}
SSL, a prevalent paradigm in natural images, has been applied to large unlabeled remote sensing images and learned useful representations for downstream tasks (classification, segmentation, and detection). However, SSL in remote sensing still has been limited and lagged behind that in natural images since the object density is different between natural and remote sensing images. Natural images, such as the general benchmark dataset ImageNet \cite{r2015imagenet}, are captured in daily scenes, and objects are mostly large in size. Hence, natural images typically have apparent foreground objects in a relatively simple background, like the fox in Figure \ref{fig:bg2M}. Remote sensing images are captured by sensors mounted on satellites and contain multiple objects in a vast and complicated scene, especially no apparent foreground objects, as illustrated in the bottom of Figure \ref{fig:bg2M}. Therefore, the high object density poses challenges to the effectiveness of the SSL paradigm in remote sensing understanding, manifesting in the following issues: (1) Mismatched positive pairs when augmenting positive pairs from one image for contrastive learning, as shown in column (a) of Figure \ref{fig:bg2M}. (2) Missing contextual information caused by random masking strategies in reconstructive learning, as exemplified in column (b) of Figure \ref{fig:bg2M}.

To address this issue, we propose CtxMIM, a novel context-enhanced self-supervised representation learning method based on reconstructive learning. As shown in Figure \ref{fig:overview}, CtxMIM designs a reconstructive branch and a context-enhanced generative branch using an asymmetric Siamese network, which can learn representations by masked image modeling. For the reconstructive branch, CtxMIM first uniformly splits an input $x$ into non-overlapping image patches $x^p$ and then randomly mask $x^p$ on patch-level to obtain a new input patches $\hat{x}^{p}$. Subsequently, CtxMIM takes as input $\hat{x}^{p}$ to reconstruct the raw pixel values of the masked patches through the encoder-decoder architecture. The reconstructive branch is detailed in the subsection \ref{subsec:Re}. Unlike natural images, the object density of remote sensing images is highly uneven, resulting in missing contextual information when randomly masking image patches. To tackle this issue, we represent the original image patches as reconstructive templates rather than directly removing the masked patches. Based on this analysis, we introduce a context-enhanced generative branch to facilitate semantic reasoning by providing meaningful contextual information during the reconstruction.

Moreover, the context-enhanced generative branch shares the encoder and the decoder with the reconstructive branch, which is detailed in the subsection \ref{subsec:Pr}. For pretraining, we minimize the distance between raw pixel values x and the reconstructive values y for reconstructive and context-enhanced branches. Additionally, a context consistency loss is proposed to guide the reconstructive process of masked patches, which is described in subsection \ref{subsec:Ob}. Finally, our proposed CtxMIM is pretrained on a large-scale unlabeled dataset and learns impressive representations, which is beneficial for various downstream tasks (\textit{e.g., image-level, object-level, and pixel-level}).

\subsection{Reconstruction of masked image patches}
\label{subsec:Re}
In the reconstruction stage, CtxMIM aims to learn local fine-grained features by restoring masked image patches. Given an input image $x \in \mathbb R^{C\times H\times W}$, CtxMIM firstly reshapes $x$ to get image sequence patches $x^p \in \mathbb R^{N\times P^2C}$, where $H, W, C$ are the image height, width, and channel, respectively, $P$ is the size of each image patch (\textit{i.e.}, the height and width), and $N=(H/P)\cdot(W/P)$ is the number of image patches. Then, a patch-wise masking operation randomly masks some image patches, and a patch embedding operation encodes each image patch to get the new sequence input $\hat{x}^{p}$. Subsequently, an encoder $f_{\theta}$ takes the sequence input $\hat{x}^{p}$ to extract latent representations $\hat{h}^{p}$, which are used in downstream tasks after pretraining. Finally, a decoder $g_{\theta}$ is used to reconstruct pixel values of masked patches $\hat{y}^{p}$ on the latent representations. In our paper, we take Swin Transformer \cite{liu2021swin} as the feature encoder $f_{swin}$ and a lightweight prediction head as the decoder $g_{mlp}$ to output raw pixel values of the masked patches by following SimMIM \cite{xie2022simmim}, which can be formulated as:
\begin{equation} \label{eq:nratio}
  \begin{split}
  \hat{y}^{p}&=g_{mlp}(f_{swin}(\hat{x}^{p}) \odot \mathbb I_{\mathcal M} (\hat{x}^p_i))
  \end{split}
\end{equation}
where $\mathbb I_{\mathcal M}(\cdot)$ is an indicator function that is 1 when each image patch $\hat{x}^p_i$ is masked, 0 otherwise. $\odot$ is an element-wise product operation.

\subsection{Context-enhanced generative branch}
\label{subsec:Pr}
Compared with natural images that are captured in daily scenes and objects are mostly with large size, remote sensing images typically cover multiple land objects with a large-scale and complex scene, which have no clear foreground objects. Therefore, some small land covers could be entirely masked, resulting in missing critical contextual information during the reconstruction. To tackle this issue, CtxMIM introduces a context-enhanced generative branch to provide contextual information. As illustrated in Figure \ref{fig:overview}, the context-enhanced generative branch inputs the image patches $x^p$ to extract latent representations $h^{p}$ with rich contextual information by the shared encoder $f_{\theta}$. Then, the latent representations $h^{p}$ are fed to the same decoder $g_{\theta}$ to predict the raw pixels $y^{p}$. The process of this branch is defined as follows:
\begin{equation} \label{eq:nratio}
  \begin{split}
  y^{p}&=g_{mlp}(f_{swin}(x^{p}) \odot \mathbb I_{\mathcal M} (x^p_i))
  \end{split}
\end{equation}
where $\mathbb I_{\mathcal M}(\cdot)$ and $\odot$ are the same operations as the reconstructive branch. Finally, the context-enhanced generative branch uses $y^{p}$ as the template to make the reconstructive branch mimic feature extraction and learning semantic extrapolating by providing meaningful contextual information.

\subsection{Training objective}
\label{subsec:Ob}
CtxMIM first realizes a reconstructive learning objective by minimizing the distance between raw pixel values $x\mid_\mathcal M$ of the masked patches and the reconstructive values $y^{Re}\mid_\mathcal M$, which is calculated as:
\begin{equation}
  {\mathcal L}_{Re} = \frac{\Vert y^{Re}\mid_\mathcal M-x\mid_\mathcal M \Vert}{N_M}
\label{equ:lrec}
\end{equation}
where $N_M$ is the number of masked pixels. $\Vert \cdot \Vert$ is a distance function for calculating similarity between $y^{Re}\mid_\mathcal M$ and $x\mid_\mathcal M$. We adopt $\ell_1$ loss for ${\mathcal L}_{Re}$ calculation in our paper.

For the context-enhanced generative branch, CtxMIM learns contextual information by calculating the predicted loss ${\mathcal L}_{Pr}$ between the predicted pixel values $Y_{Pr}$ and raw pixel values $X$, which is similar to Equation \ref{equ:lrec}. Then, a context consistency loss ${\mathcal L}_{Cc}$ maximizes the similarity between $y^{Re}\mid_\mathcal M$ and the corresponding portion $y^{Pr}\mid_\mathcal M$, which is defined as:
\begin{equation}
  {\mathcal L}_{Cc} = \text{Dist}(y^{Re}\mid_\mathcal M-y^{Pr}\mid_\mathcal M)
\label{equ:lcon}
\end{equation}
where $\text{Dist}(\cdot)$ is a similarity function to guide CtxMIM learning good contextual information. In our paper, we adopt $\ell_1$ loss in the $\text{Dist}(\cdot)$ function. The context consistency loss ${\mathcal L}_{Cc}$ mitigates pretraining convergence instability due to informational deficiencies.

Finally, the joint training objective of CtxMIM is calculated by the Equation \ref{equ:overloss}:
\begin{equation}
  {\mathcal L} = {\mathcal L}_{Re} + {\mathcal L}_{Pr} + {\mathcal L}_{Cc}
\label{equ:overloss}
\end{equation}
For ${\mathcal L}_{Re}$ and ${\mathcal L}_{Pr}$, the gradients are propagated backward along their respective branches. As shown in Figure \ref{fig:overview}, the gradients from ${\mathcal L}_{Cc}$ are propagated backward only to the reconstructive branch, prompting the encoder to mimic feature extraction and learning semantic extrapolating by providing meaningful contextual information. Additionally, it can also avoid trivial constant solutions by a stop-gradient operation. Finally, CtxMIM learns representations with rich semantic and local information by pretraining in a multi-task learning manner.

The proposed CtxMIM adopts Swin-B as the encoder and is implemented by the PyTorch framework. We use an AdamW optimizer with hyper-parameters $\beta_1=0.9$, $\beta_2=0.999$, and $\epsilon=1\times10^{-8}$, and the learning rate is set to $1\times10^{-5}$. The training lasts 200 epochs with 8 NVIDIA Tesla V100 GPUs.
For pretraining, we collect a large-scale unlabeled remote sensing dataset captured from WorldView-3 using Google Earth Engine \cite{gorelick2017google}. This pretraining dataset is characterized by diverse topography and terrain, covering most regions (cities, villages, rivers, mountains, woodlands, and other terrain) of Asia.

\section{Experiments}
\label{sec:exp}
This section describes datasets and experiment settings for transfer learning. We evaluate our model on four downstream tasks and compare its performance with SOTA methods quantitatively and qualitatively. Finally, we conduct extensive ablation studies to verify the effectiveness of the proposed pretraining framework.

\subsection{Experimental Settings}
In downstream experiments, we select land cover classification, semantic segmentation, object detection, and instance segmentation, covering image-level, object-level, and pixel-level remote sensing tasks. To comprehensively confirm the performance of our method, we compare CtxMIM with supervised learning and SOTA self-supervised learning models, which are downloaded from their public websites. Finally, extensive experiments show that CtxMIM has strong performance on downstream tasks.

\noindent \textbf{Sup.} We employ three general backbones (ResNet-50 \cite{he2016resnet}, ViT \cite{dosovitskiy2020vit}, and Swin Transformer \cite{liu2021swin}) initialized with ImageNet pretraining for supervised learning comparison. In experiments, ``Random Init.'' and ``ImageNet Pre.'' denote that different backbones are initialized with random initialization and ImageNet pretraining, respectively. Besides, SatLas \cite{bastani2023satlas} is compared as a remote sensing supervised method since it is pretrained on a large-scale remote sensing dataset with 137 categories under seven label types.

\noindent \textbf{Self-Sup.} For self-supervised learning comparison, we evaluate our approach with contrastive and reconstructive methods. We present additional details regarding the comparison methods in terms of datasets and experiment settings in the following:

\begin{enumerate}
\item \textbf{SeCo} \cite{m2021seco}, based on Momentum Contrast (MoCov2) \cite{c2020mocov2}, pretrains a ResNet-50 backbone for 200 epochs with a batch size of 256. It collects a new dataset from Sentinel-2 \cite{drusch2012sentinel}, designed to capture seasonal changes by sampling five images from distinct dates.

\item \textbf{Geo-Aware} \cite{a2021geoaware} pretrains a ResNet-50 backbone based on MoCov2 \cite{c2020mocov2} for 200 epochs with a batch size of 256. It creates GeoImageNet by extracting geo-coordinates for ImageNet \cite{r2015imagenet}. Finally, Geo-Aware uses the Functional Map of the World (fMoW) \cite{christie2018fmow} and GeoImageNet for pretraining.

\item \textbf{CACo} \cite{mall2023caco} adopts the same pretraining settings and backbone as SeCo \cite{m2021seco}. CACo reconsiders the temporal signal to contrast images with long-term and short-term differences, spanning four years for the long-term and one year for the short-term.

\item \textbf{TOV} \cite{tao2023tov} is the original vision model with a ResNet50 backbone in the remote sensing field. By analyzing the influences of different data sampling and the selection of learning paths, it releases a benchmark dataset for pretraining. Finally, TOV is pretrained for 800 epochs with a batch size of 1024.

\item \textbf{CMID} \cite{muhtar2023cmid} is a unified SSL framework which learns representations by combining the CL and MIM. CMID pretrains ResNet-50 and Swin-B backbones for 200 epochs on the MillionAID dataset with a batch size of 512 and 256, respectively.

\item \textbf{GeRSP} \cite{huang2024gersp} is a MoCo-based framework using a teacher-student architecture. It pretrains a ResNet-50 backbone on the unlabeled Million-AID (MAID) \cite{long2021maid} and ImageNet \cite{r2015imagenet} for 200 epochs with a batch size of 128.

\item \textbf{ScaleMAE} \cite{reed2023scalemae} is based on the MAE architecture \cite{he2022mae}, employing ViT-Large as the backbone. It uses the Functional Map of the World (FMoW) \cite{christie2018fmow} RGB training set for 800 epochs with a batch size of 128.

\item \textbf{SatMAE++} \cite{noman2024satmaepp} is a multi-scale pre-training approach, employing a Mask Autoencoder (MAE) framework. SatMAE++ pretrains a ViT-Large \cite{dosovitskiy2020vit} model on the Functional Map of the World (FMoW) \cite{christie2018fmow} RGB training set. It lasts for 800 epochs with a batch size of 64.

\begin{table*}
\centering
\caption{Top-1 accuracy on EuroSAT and NWPU-RESISC45 for land cover classification. CtxMIM obtains the best performance.}
\begin{tabular}{lll|c|c}
\hline
~ & \multirow{2}*{\makecell[c]{Method}} & \multirow{2}*{\makecell[c]{Backbone}} & EuroSAT & NWPU-RESISC45 \\
~ &                               ~     &           ~      & Top-1 Acc. $\uparrow$ & Top-1 Acc. $\uparrow$ \\
\hline
\multirow{7}*{Sup.} & Random Init.    & ResNet50 & 77.74 & 45.88 \\
~ & ImageNet Pre.                     & ResNet50 & 98.38 & 93.88 \\
~ & ImageNet Pre.                     & ViT-B    & 89.83 & 82.68 \\
~ & ImageNet Pre.                     & ViT-L    & 89.94 & 84.36 \\
~ & ImageNet Pre.                     & Swin-B   & 93.38 & 90.80 \\
\cline{2-5}
~ & SatLas \cite{bastani2023satlas}                  & ResNet50 & 96.62 & 86.23         \\
~ & SatLas \cite{bastani2023satlas}                  & Swin-B & 83.44   & 79.88         \\
\hline
\hline
\multirow{13}*{Self-Sup.} & SeCo \cite{m2021seco}    & ResNet50 & 95.09 & 61.52         \\
~ & Geo-Aware \cite{a2021geoaware}                   & ResNet50 & 95.66 & 81.15         \\
~ & CACo \cite{mall2023caco}                         & ResNet50 & 96.59 & 68.60         \\ 
~ & TOV \cite{tao2023tov}                            & ResNet50 & 57.81 & 46.74         \\ 
~ & CMID \cite{muhtar2023cmid}                       & ResNet50 & 91.07 & 76.07         \\
~ & GeRSP \cite{huang2024gersp}                      & ResNet50 & 98.29 & 92.83         \\
\cline{2-5}
~ & ScaleMAE \cite{reed2023scalemae}                 & ViT-L  & 86.12   & 78.04        \\
~ & SatMAE++ \cite{noman2024satmaepp}                & ViT-L  & 87.83   & 82.60         \\
~ & CrossScaleMAE \cite{tang2024csmae}               & ViT-L  & 72.81   & 84.76         \\
~ & SpectralGPT \cite{hong2024spectralgpt}           & ViT-B  & 84.61   & 71.04         \\
~ & GFM \cite{mendieta2023gfm}                       & Swin-B & 88.87   & 84.42         \\
~ & CMID \cite{muhtar2023cmid}                       & Swin-B & 92.72   & 93.23         \\
~ & CtxMIM (ours)                                     & Swin-B & \textbf{98.87}   & \textbf{95.20}  \\
\hline
\end{tabular}
\label{tab:mmcls}
\end{table*}

\item \textbf{CrossScaleMAE} \cite{tang2024csmae}, based on Masked Auto-Encoder (MAE), pretrains a ViT-Large backbone on the Functional Map of the World (fMoW) \cite{christie2018fmow} RGB training set. It is pretrained for 400 epochs with a batch size of 512.

\item \textbf{SpectralGPT} \cite{hong2024spectralgpt} is a foundation model by considering unique characteristics of spectral data in a MAE framework. It pretrains a ViT-Base backbone on an extensive dataset from the Sentinel2 satellite with over one million images. Finally, SpectralGPT is pretrained for 200 epochs with a batch size of 16.

\item \textbf{GFM} \cite{mendieta2023gfm} collects a compact yet diverse dataset GeoPile from multiple sources to promote feature diversity. GFM introduces a multi-objective continual pretraining paradigm based on MIM \cite{xie2022simmim}, which can leverage the representations of ImageNet to learn valuable features. Finally, it pretrains a Swin-B backbone for 100 epochs with a batch size of 128.
\end{enumerate}

\subsection{Land Cover Classification}
\noindent \textbf{Implementation Details}.
We take two land cover classification datasets: (1) EuroSAT \cite{helber2019eurosat} is captured by the Sentinel-2 satellite over European cities. This dataset has 27,000 labeled images with a size of 64$\times$64. It contains 10 classes with 2,000-3,000 images per class. Following SeCo \cite{m2021seco}, we adopt the official train/val splits on EuroSAT. (2) NWPU-RESISC45 \cite{cheng2017nwpu} is a publicly available benchmark for remote sensing classification, which is published by Northwestern Polytechnical University (NWPU). The dataset contains 31,500 images with spatial resolutions ranging from 30 to 0.2m and a size of 256$\times$256. It covers 45 categories with 700 images in each category. Following GeRSP \cite{huang2024gersp}, we use the same train/val splits for this experiment.

For the land cover classification task, we initialize a backbone with different ed models and fine-tune a linear classifier head with ground truth labels in a supervised manner. We use an SGD optimizer with a learning rate of $1\times10^{-4}$, a momentum of 0.9, and a weight decay of 0.0005. All the models are trained for 100 epochs with a batch size of 64. During training, we adopt a linear-step policy with a decay ratio of 10 at 60\% and 80\% of the epochs. Finally, we report the top-1 accuracy for performance comparison.

\noindent \textbf{Quantitative Results}.
Table \ref{tab:mmcls} shows the top-1 accuracy of different methods on two datasets using fully supervised and self-supervised learning.
For the EuroSAT dataset, we can see that CtxMIM outperforms the ImageNet supervised Swin-B backbone by +5.49\%, illustrating that SSL on a large-scale remote sensing dataset improves performance for remote sensing downstream tasks. Additionally, CtxMIM achieves +6.15\% higher top-1 accuracy than reconstructive learning methods, demonstrating that our approach learns robust and good transferable remote sensing representations for the land cover classification task.
For the NWPU-RESISC45 dataset, we observe that CtxMIM improves the model of the Swin-B backbone initialized with the ImageNet supervised pretraining by +4.4\% higher, proving that the domain gap exists between natural and remote sensing images. We find that CtxMIM outperforms contrastive and reconstructive pretraining methods by +2.37\% and +1.97\% higher, respectively.

\begin{table*}
\centering
\caption{Mean intersection over union (mIoU) and mean accuracy (mAcc) on SpaceNet (Rio) semantic segmentation task. Our method consistently improves supervised and self-supervised learning by large margins.}
\begin{tabular}{lll|cc}
\hline
~ & Method & Backbone & mIoU $\uparrow$ & mAcc $\uparrow$ \\
\hline
\multirow{7}*{Sup.} & Random Init. & ResNet50         & 72.24 & 80.20 \\
~ & ImageNet Pre.                  & ResNet50         & 75.70 & 83.98 \\ 
~ & ImageNet Pre.                  & ViT-B            & 75.47 & 83.78 \\
~ & ImageNet Pre.                  & ViT-L            & 76.34 & 84.34 \\
~ & ImageNet Pre.                  & Swin-B           & 72.56 & 81.21 \\
\cline{2-5}
~ & SatLas \cite{bastani2023satlas}                  & ResNet50  & 74.56 & 82.38          \\
~ & SatLas \cite{bastani2023satlas}                  & Swin-B    & 67.86 & 76.22          \\
\hline
\hline
\multirow{13}*{Self-Sup.} & SeCo \cite{m2021seco}    & ResNet50  & 73.89 & 81.63          \\
~ & Geo-Aware \cite{a2021geoaware}                   & ResNet50  & 74.99 & 82.30          \\
~ & CACo \cite{mall2023caco}                         & ResNet50  & 74.98 & 82.77          \\
~ & TOV \cite{tao2023tov}                            & ResNet50  & 74.12 & 81.89          \\
~ & CMID \cite{muhtar2023cmid}                       & ResNet50  & 71.53 & 79.28          \\
~ & GeRSP \cite{huang2024gersp}                      & ResNet50  & 76.29 & 84.47          \\
\cline{2-5}
~ & ScaleMAE \cite{reed2023scalemae}                 & ViT-L     & 77.21 & 85.12          \\
~ & SatMAE++ \cite{noman2024satmaepp}                & ViT-L     & 77.25 & 85.40          \\
~ & CrossScaleMAE \cite{tang2024csmae}               & ViT-L     & 75.55 & 83.87          \\
~ & SpectralGPT \cite{hong2024spectralgpt}           & ViT-B     & 69.17 & 77.78          \\
~ & GFM \cite{mendieta2023gfm}                       & Swin-B    & 72.81 & 81.71           \\
~ & CMID \cite{muhtar2023cmid}                       & Swin-B    & 65.98 & 73.72          \\
~ & CtxMIM (ours)                                     & Swin-B    & \textbf{79.22} & \textbf{87.20} \\
\hline
\end{tabular}
\label{tab:spacenetrio}
\end{table*}

\begin{table*}
\centering
\caption{Average precision (AP) and average recall (AR) under different intersection over union thresholds on DOTA object detection task. $F1$/$F1_{75}$ calculated by $AP$/$AP_{75}$ and $AR$/$AR_{75}$ measures the performance more comprehensively/accurately.}
\begin{tabular}{llcccccccccccc|cc}
\hline
Method  & Backbone & AP & $AP_{50}$ & $AP_{75}$ & $AP_s$ & $AP_m$ & $AP_l$ & AR & $AR_{50}$ & $AR_{75}$ & $AR_s$ & $AR_m$ & $AR_l$ & $F1$ & $F1_{75}$ \\
\hline
Random Init.                            & ResNet50 & 24.2 & 43.9 & 25.0 & 13.4 & 25.7 & 27.5 & 40.1 & 70.8 & 40.9 & 22.7 & 43.8 & 52.5 & 30.2 & 31.0 \\
ImageNet Pre.                           & ResNet50 & 35.9 & 61.3 & 38.1 & 21.2 & 37.2 & 45.6 & 46.7 & 75.2 & 50.4 & 29.1 & 50.2 & 59.5 & 40.6 & 43.4 \\
ImageNet Pre.                           & Swin-B & 36.4 & \textbf{64.3} & 37.8 & \textbf{23.3} & \textbf{38.9} & 44.1 & 48.7 & \textbf{81.9} & 51.8 & \textbf{32.2} & 53.6 & 59.4 & 41.7 & 43.7 \\ 
\hline
SatLas    \cite{bastani2023satlas}     & ResNet50 & 30.0 & 52.8 & 31.5 & 15.8 & 30.7 & 38.7 & 41.1 & 69.4 & 43.7 & 22.9 & 44.0 & 54.0 & 34.7 & 36.6 \\
SatLas    \cite{bastani2023satlas}     & Swin-B & 27.8 & 50.7 & 28.7 & 18.0 & 28.6 & 34.0 & 42.5 & 75.1 & 43.8 & 27.7 & 46.9 & 52.7 & 33.6 & 34.7 \\
\hline
\hline
SeCo      \cite{m2021seco}             & ResNet50 & 32.0 & 55.4 & 33.9 & 18.6 & 32.7 & 38.5 & 44.7 & 74.7 & 47.6 & 28.0 & 48.1 & 55.1 & 37.3 & 39.6 \\
Geo-Aware \cite{a2021geoaware}         & ResNet50 & 33.2 & 58.0 & 34.3 & 17.8 & 34.4 & 41.9 & 44.5 & 73.9 & 47.4 & 26.7 & 47.9 & 56.1 & 38.0 & 39.8 \\
CACo      \cite{mall2023caco}          & ResNet50 & 33.7 & 58.5 & 35.4 & 19.9 & 34.6 & 41.4 & 45.7 & 75.5 & 48.5 & 28.8 & 48.8 & 56.8 & 38.8 & 40.9 \\
TOV       \cite{tao2023tov}            & ResNet50 & 35.2 & 61.7 & 36.6 & 21.6 & 36.2 & 44.2 & 46.4 & 76.8 & 49.3 & 30.1 & 49.3 & 58.5 & 40.0 & 42.0 \\
CMID      \cite{muhtar2023cmid}        & ResNet50 & 36.3 & 63.0 & 38.3 & 21.1 & 38.0 & 44.8 & 48.1 & 79.9 & 51.8 & 28.8 & 52.9 & 60.0 & 41.4 & 44.0 \\ 
GeRSP     \cite{huang2024gersp}        & ResNet50 & 37.6 & 63.3 & 39.4 & 22.0 & 38.6 & 47.3 & 47.9 & 76.0 & 51.9 & 29.8 & 51.4 & 60.2 & 42.1 & 44.8 \\
\hline
GFM    \cite{mendieta2023gfm}          & Swin-B & 35.4 & 62.7 & 36.8 & 22.5 & 38.0 & 44.1 & 48.8 & 81.4 & 52.1 & 32.0 & 53.5 & 60.1 & 41.0 & 43.1 \\
CMID   \cite{muhtar2023cmid}           & Swin-B & 27.3 & 51.0 & 27.5 & 17.2 & 28.7 & 32.6 & 42.2 & 75.9 & 43.1 & 26.4 & 47.0 & 52.0 & 33.2 & 33.6 \\
CtxMIM (ours)                           & Swin-B & \textbf{37.9} & 62.8 & \textbf{40.9} & 22.9 & 38.6 & \textbf{48.8} & \textbf{49.9} & 78.6 & \textbf{54.3} & 31.9 & \textbf{53.8} & \textbf{62.8} & \textbf{43.1} & \textbf{46.7} \\ 
\hline
\end{tabular}
\label{tab:dota}
\end{table*}
\subsection{Semantic Segmentation}
\noindent \textbf{Implementation Details}.
We use the SpaceNet (Rio) dataset \cite{van2018spacenet} for the semantic segmentation task. This dataset contains 6,940 satellite images with binary building masks. We fine-tune a PSANet \cite{zhao2018psanet} with the pretrained Swin-B backbone by CtxMIM. We use an AdamW optimizer with a batch size of 8 and a learning rate of $1\times10^{-4}$. We report mean intersection over union (mIoU) and mean accuracy (mAcc) to compare CtxMIM with comparison methods.

\noindent \textbf{Quantitative Results}.
Table \ref{tab:spacenetrio} presents mIoU and mAcc values of different methods on SpaceNet (Rio). The results show that our method achieves more considerable performance gains by +6.66\% and +5.99\% for supervised learning from the ImageNet pretrained weight with a Swin-B backbone. Compared with self-supervised learning, our method improves the mIoU performance by +2.93\% and +1.97\% over SOTA contrastive and reconstructive learning methods. Consistent experimental results demonstrate the effectiveness of our method for the semantic segmentation task.

\subsection{Object Detection}
\noindent \textbf{Implementation Details}.
We use the DOTA dataset \cite{xia2018dota} to verify the object detection performance. DOTA dataset consists of 2,806 high-resolution satellite images with sizes ranging from 800$\times$800 to 4,000$\times$4,000. This dataset has 15 different categories with bounding box annotations and is split into 800$\times$800 following GeRSP \cite{huang2024gersp}. Then, we fine-tune a RetinaNet \cite{lin2017focal} with an AdamW optimizer, a batch size of 4, a learning rate of $1\times10^{-5}$, and a weight decay of 0.05. We use the standard MS-COCO metrics \cite{lin2014mscoco}, including average precision (AP) and average recall (AR) under different intersection over union thresholds, to compare CtxMIM with supervised and self-supervised methods on this task. Moreover, we calculate $F1$ and $F1_{75}$ to evaluate CtxMIM more comprehensively and accurately.

\noindent \textbf{Quantitative Results}.
Table \ref{tab:dota} reports results from comprehensive experiments on the DOTA dataset compared with supervised and self-supervised learning methods. CtxMIM consistently obtains the best performance across supervised and self-supervised learning methods. Specifically, CtxMIM gains +1.4\% and +3\% improvements on $F1$ and $F1_{75}$ over ImageNet supervised learning with a Swin-B backbone. Furthermore, CtxMIM improves the indicator $F1$ by +1\% and 2.1\% compared with contrastive and reconstructive learning methods. The extraordinary performance demonstrates that CtxMIM learns impressive representations for the object detection task.

\begin{table*}
\centering
\caption{$AP^{m}$, $AP_{50}^{m}$, and $AP_{75}^{m}$ on SpaceNet (Las Vegas) instance segmentation task. CtxMIM outperforms comparison methods (+2.2\% and +1.1\% on $AP^m_{75}$ for Sup. and Self-Sup., respectively).}
\begin{tabular}{lllccc}
\hline
~ & Method & Backbone & $AP^{m}$ & $AP_{50}^{m}$ & $AP_{75}^{m}$ \\
\hline
\multirow{7}*{Sup.} & Random Init.  & ResNet50 & 16.3 & 34.8 & 13.1 \\
~ & ImageNet Pre.                   & ResNet50 & 51.0 & 82.6 & 60.3 \\
~ & ImageNet Pre.                   & ViT-B    & 58.4 & 88.1 & 69.7 \\
~ & ImageNet Pre.                   & ViT-L    & 58.8 & 88.2 & 70.7 \\
~ & ImageNet Pre.                   & Swin-B   & 45.6 & 78.4 & 50.8 \\
\cline{2-6}
~ & SatLas \cite{bastani2023satlas}               & ResNet50 & 39.3 & 73.4 & 39.3 \\
~ & SatLas \cite{bastani2023satlas}               & Swin-B   & 29.8 & 59.2 & 27.3 \\
\hline
\hline
\multirow{13}*{Self-Sup.} & SeCo \cite{m2021seco} & ResNet50 & 40.8 & 73.7 & 42.4  \\
~ & Geo-Aware \cite{a2021geoaware}                & ResNet50 & 48.9 & 80.9 & 56.3  \\
~ & CACo \cite{mall2023caco}                      & ResNet50 & 44.9 & 77.8 & 49.4  \\
~ & TOV \cite{tao2023tov}                         & ResNet50 & 41.5 & 74.2 & 43.8  \\ 
~ & CMID \cite{muhtar2023cmid}                    & ResNet50 & 42.9 & 74.1 & 47.8  \\
~ & GeRSP \cite{huang2024gersp}                   & ResNet50 & 54.3 & 85.1 & 65.4  \\
\cline{2-6}
~ & ScaleMAE \cite{reed2023scalemae}              & ViT-L    & 60.0 & \textbf{89.4} & 71.8 \\
~ & SatMAE++ \cite{noman2024satmaepp}             & ViT-L    & 59.4 & 89.0 & 71.7  \\
~ & CrossScaleMAE \cite{tang2024csmae}            & ViT-L    & 58.3 & 87.5 & 70.0  \\
~ & SpectralGPT \cite{hong2024spectralgpt}        & ViT-B    & 57.9 & 88.0 & 69.2  \\
~ & GFM \cite{mendieta2023gfm}                    & Swin-B   & 47.2 & 80.0 & 53.5  \\
~ & CMID \cite{muhtar2023cmid}                    & Swin-B   & 32.7 & 63.6 & 31.3  \\
~ & CtxMIM (ours)                  & Swin-B & \textbf{60.5} & 87.8 & \textbf{72.9}  \\
\hline
\end{tabular}
\label{tab:spacenetLV}
\end{table*}

\begin{table}
\centering
\caption{Ablation study. LcC, SS, OD, and IS denote land cover classification, semantic segmentation, object detection, and instance segmentation, respectively. `- C-E' removes the context-enhanced generative branch from CtxMIM.}
\begin{tabular}{l|c|c|c|c}
\hline
\multirow{2}*{\makecell[c]{Method}} & LcC & SS & OD & IS   \\
~        & (Acc.) & (mIoU) & ($F1_{75}$)  & ($AP^m$)            \\
\hline
CtxMIM     & $\textbf{98.87}$ & $\textbf{79.22}$ & $\textbf{46.70}$  & $\textbf{60.50}$ \\
- C-E      & 96.90            & 76.91            & 44.0             & 59.10         \\
\hline
\end{tabular}
\label{tab:ablation}
\end{table}

\subsection{Instance Segmentation}
\noindent \textbf{Implementation Details}.
SpaceNet (LasVegas) dataset \cite{van2018spacenet} is used for instance segmentation task, consisting of 3,851 images across Las Vegas collected from WorldView-3 satellites. The size of images is 650$\times$650 with a spatial resolution of 30 cm/pixel. The dataset contains 151,367 building polygon footprints in GeoJSON format. In our experiment, this dataset is randomly divided into train/test/validation subsets with a ratio of 8:1:1. We fine-tune MaskRCNN \cite{he2017mask} with the ed Swin-B backbone from our CtxMIM on SpaceNet (LasVegas), using an AdamW optimizer, a learning rate of 0.0001, a batch size of 4. Finally, we report $AP^m$, $AP_{50}^{m}$, and $AP_{75}^{m}$ under different intersections over union thresholds in the segmentation level.

\noindent \textbf{Quantitative Results}.
As shown in Table \ref{tab:spacenetLV}, our approach learns transferable representations, outperforming all the comparison methods. Compared with supervised learning, CtxMIM improves the ImageNet supervised pretraining by +1.7\% $AP^{m}$ and +2.2\% $AP_{75}^{m}$, respectively. Moreover, CtxMIM achieves +0.5\% $AP^{m}$ and +1.1\% $AP_{75}^{m}$ higher compared with the SOTA self-supervised learning method ScaleMAE \cite{reed2023scalemae}. The impressive results illustrate that CtxMIM learns good representations with high-level semantic and fine-grained details.

\begin{figure*} [htbp]
\centering
\includegraphics[width=1.0\linewidth]{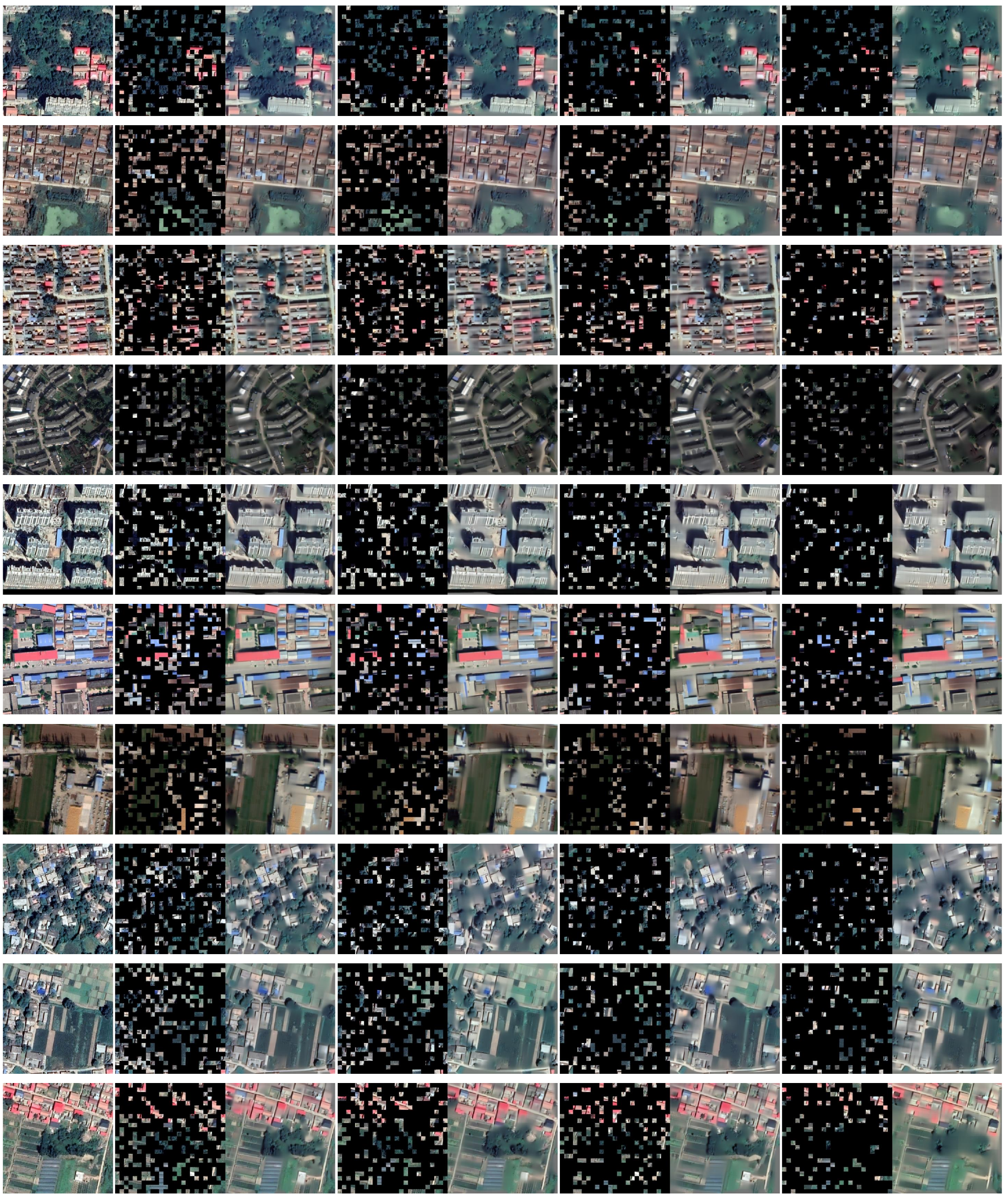}
\caption{Some CtxMIM reconstructive samples on our pretraining dataset. Each row shows the ground truth, four 2-tuples (the masked image and the CtxMIM reconstruction) of different masking ratios 70\%, 75\%, 80\%, and 85\% from left to right.}
\label{fig:recocndata}
\end{figure*}

\begin{figure*} [htbp]
\centering
\includegraphics[width=1.0\linewidth]{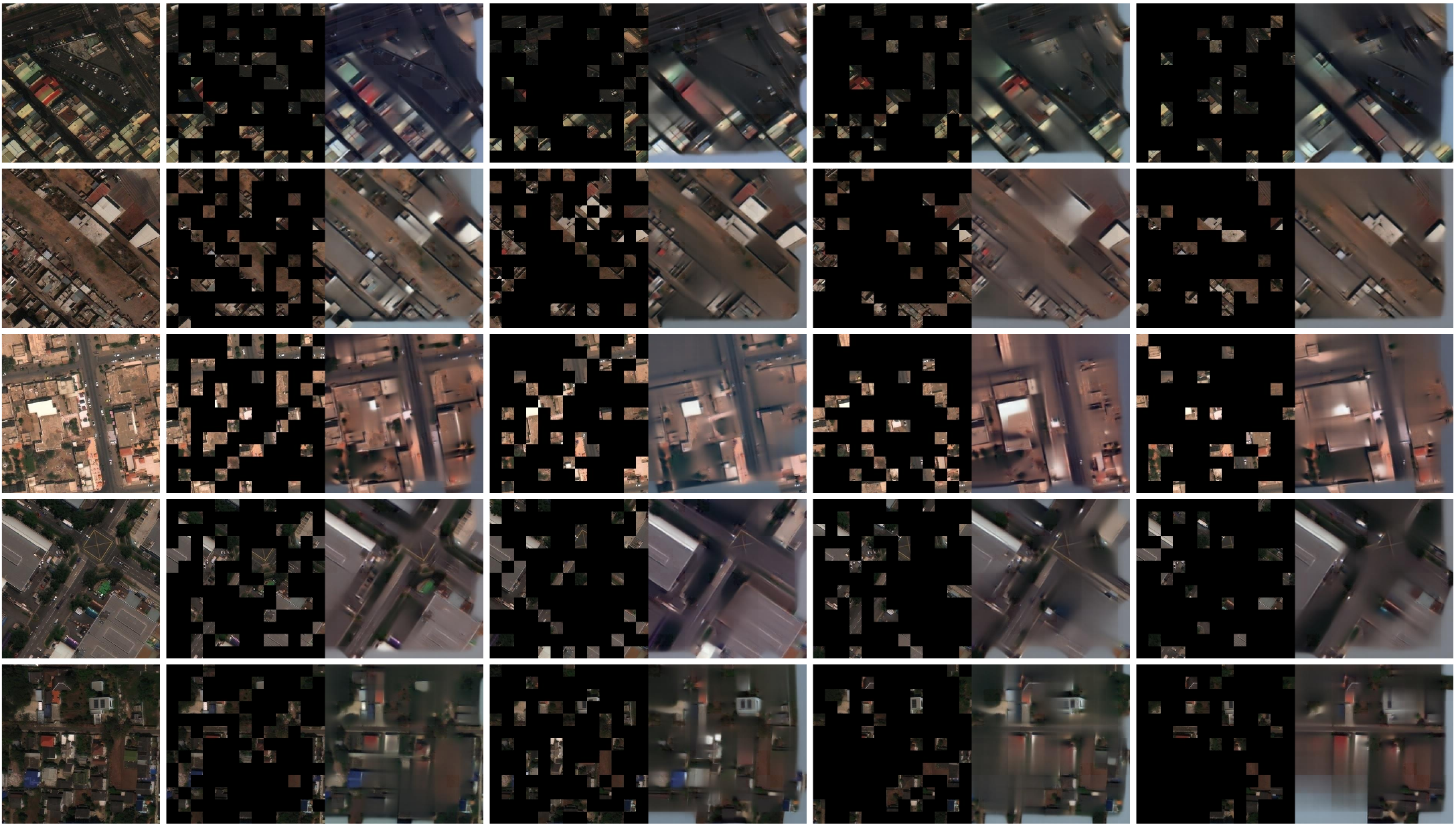}
\includegraphics[width=1.0\linewidth]{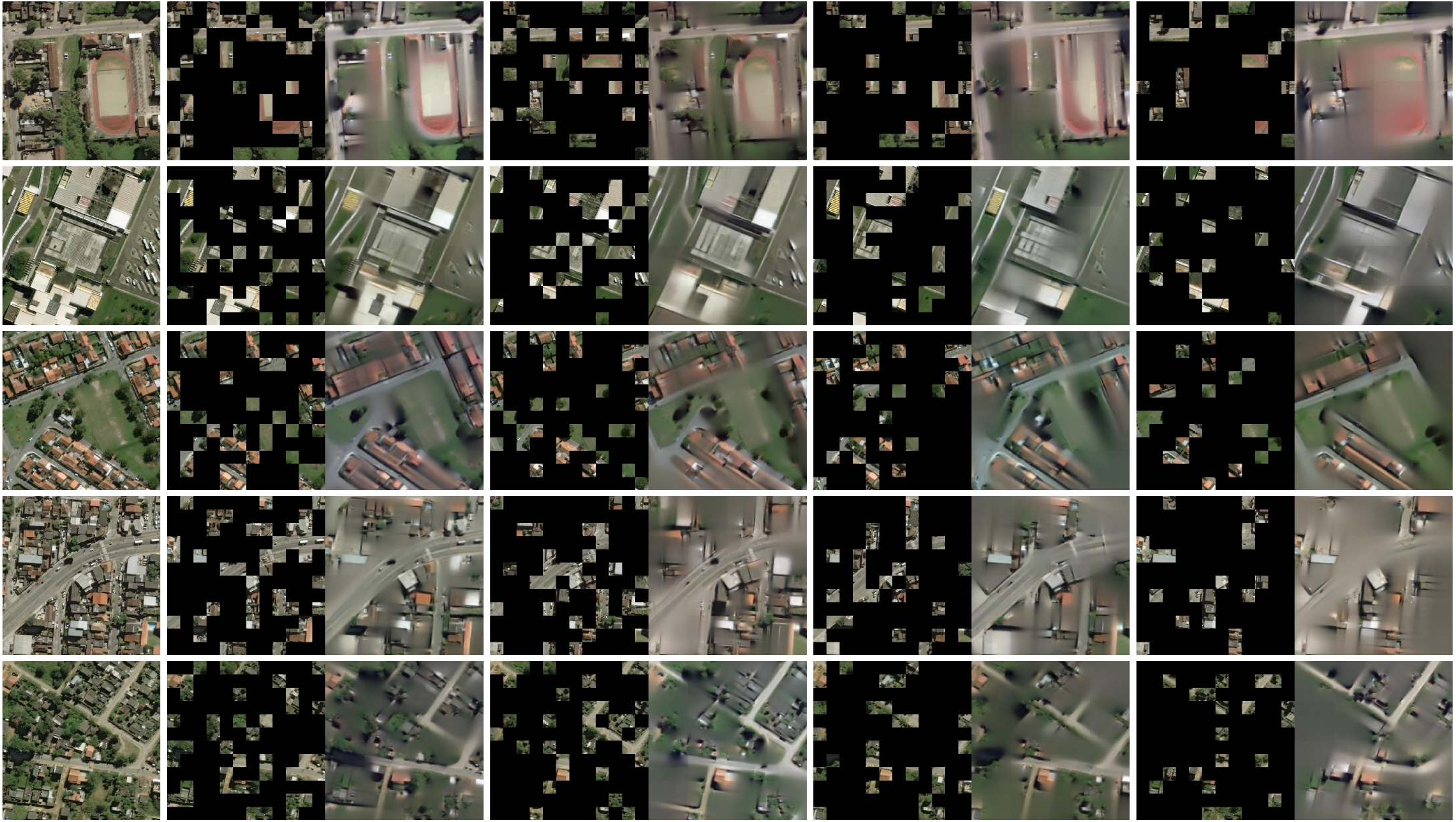}
\caption{Some CtxMIM reconstructive samples on xView \cite{lam2018xview} and SpaceNet \cite{van2018spacenet}. Each row shows the ground truth, four 2-tuples (the masked image and the CtxMIM reconstruction) of different masking ratios 70\%, 75\%, 80\%, and 85\% from left to right.}
\label{fig:recoxview}
\end{figure*}
\subsection{Ablation Study}
\label{subsec:ablation}
We conduct ablation studies to verify the effectiveness of the designed context-enhanced generative branch on four typical remote sensing tasks, respectively. We adopt general datasets and critical indicators to measure performance, including EuroSAT \cite{helber2019eurosat}, SpaceNet (Rio) \cite{van2018spacenet}, DOTA \cite{xia2018dota}, and SpaceNet (LasVegas) \cite{van2018spacenet}. As reported in Table \ref{tab:ablation}, CtxMIM outperforms CtxMIM without the context-enhanced generative branch by +1.97\% on top-1 accuracy, 2.31\% on mIoU, +2.7\% on $F1_{75}$, and +1.4\% on $AP^m$. The consistent improvements have confirmed that the context-enhanced generative branch can significantly improve the performance, illustrating its superiority.

Additionally, Figures \ref{fig:recocndata} and \ref{fig:recoxview} show some CtxMIM reconstructive samples on various datasets with different masking ratios 70\%, 75\%, 80\%, and 85\%. The qualitative results demonstrate CtxMIM's proficiency in recovering high-quality masked patches, particularly at higher masking ratios. This observation demonstrates its ability to acquire robust features for remote sensing image analysis.

\section{Discussion}
\label{sec:dis}
In this section, we further analyze our method in terms of model structure and performance and then discuss limitations and future work about SSL in remote sensing.

\noindent \textbf{Model scalability}.
SSL methods in remote sensing employ contrastive learning to pretrain foundation models and have demonstrated higher performance on different remote sensing tasks. For extensive data augmentations, these methods consider intrinsic characteristics of remote sensing images to obtain contrastive pairs, like SeCo \cite{m2021seco}, CACo \cite{mall2023caco}, and SatMAE \cite{c2022satmae} leveraging the temporal information, and Geo-Aware \cite{a2021geoaware} leveraging geo-location information. However, the well-designed handcrafted data structure is a double-edged sword that limits the capabilities of foundation models to some extent. Besides, remote sensing image pairs are sensitive to significant changes in the temporal span, such as major urban development \cite{mall2023caco}, natural disasters, and the external environment. Hence, CtxMIM designs a novel MIM framework by taking the natural advantage of abundant remote sensing images captured by sensors daily, making the foundation model easily extended to a larger scale in further research.

\noindent \textbf{Exploring more effective SSL paradigm}.
Self-supervised learning (SSL) has exhibited remarkable performance in remote sensing analysis. However, SSL in remote sensing lags behind natural images, primarily due to the object density gap between the two domains. As illustrated in Figure \ref{fig:bg2M}, natural images, captured in daily scenes, typically feature apparent foreground objects against a relatively simple background. In contrast, remote sensing images, acquired by sensors on satellites, encompass multiple objects within vast and complex scenes, especially no apparent foreground objects. Due to the enormous object density gap, advanced SSL methods in remote sensing still suffer from \textbf{3M} problem: (1) \textbf{M}ismatched positive pairs augmented from one remote sensing image; (2) \textbf{M}issing contextual information caused by random masking strategies; (3) \textbf{M}echanism in pretraining dataset collection. We rethink: \textit{is SSL paradigm effective for remote sensing representation learning}? We explore this answer from the characteristics of remote sensing data and tasks. Since most remote sensing tasks are dense predictions at the object or pixel levels, contrastive method learning representations from two augmented views is less optimal for remote sensing analysis. Our CtxMIM leverages the original image patches masked in MIM to learn good representations from extensive remote sensing images and has illustrated its performance on various tasks.

\noindent \textbf{Limitations and future work}.
Although CtxMIM is a simple yet powerful MIM method for remote sensing pretraining, we argue that collaborative learning by combining contrastive and reconstructive learning can yield robust representations for remote sensing analysis.

In future work, we will explore a new paradigm of collaborative learning, design a more efficient mask strategy, and employ patch pairs of masked and original to jointly embed contrastive learning into reconstructive learning. With the simple pipeline and the solid performance, we hope this work will inspire other works, not just in the remote sensing community but also in the industry and medical image analysis community.

\section{Conclusion}
\label{sec:con}
This paper presents CtxMIM, a simple yet efficient masked image modeling for remote sensing representation learning. CtxMIM is built on the insight that the high object density in remote sensing images significantly impedes semantic reasoning for reconstructive learning due to missing contextual information. Therefore, CtxMIM utilizes the original patches and introduces a context-enhanced generative branch, which can provide contextual information through the context consistency constraint. Finally, we train CtxMIM by collecting a large-scale unlabeled remote sensing dataset with 1.28 million images covering rich topography and terrain. Comprehensive experiments on land cover classification, semantic segmentation, object detection, and instance segmentation demonstrate that CtxMIM learns good features with high generalization and transferable ability, significantly outperforming supervised and self-supervised methods.

\bibliographystyle{IEEEtran}
\bibliography{paper}

\end{document}